\documentclass{article}

\usepackage{microtype}
\usepackage{graphicx}
\usepackage{subfigure}
\usepackage{booktabs} 

\usepackage{hyperref}

\usepackage[accepted]{icml2020}

\usepackage[english]{babel}
\usepackage[autostyle, english = american]{csquotes}
\MakeOuterQuote{"}
\usepackage{amssymb}
\usepackage{amsmath}
\usepackage{amsthm}
\usepackage{float} 
\DeclareMathOperator*{\argmin}{argmin}
\DeclareMathOperator*{\argmax}{argmax}
\newcommand{\norm}[1]{\left\lVert#1\right\rVert}

\def\XXint#1#2#3{{\setbox0=\hbox{$#1{#2#3}{\int}$ }
		\vcenter{\hbox{$#2#3$ }}\kern-.6\wd0}}

\begin{document}

\twocolumn[
\icmltitle{Projective Preferential Bayesian Optimization}




\begin{icmlauthorlist}
\icmlauthor{Petrus Mikkola}{aalto_cs}
\icmlauthor{Milica Todorovi\'{c}}{aalto_phy}
\icmlauthor{Jari J{\"a}rvi}{aalto_phy}
\icmlauthor{Patrick Rinke}{aalto_phy}
\icmlauthor{Samuel Kaski}{aalto_cs,manchester}
\end{icmlauthorlist}

\icmlaffiliation{aalto_cs}{Helsinki Institute for Information Technology HIIT, Department of Computer Science, Aalto University, Espoo, Finland}
\icmlaffiliation{aalto_phy}{Department of Applied Physics, Aalto University, Espoo, Finland}
\icmlaffiliation{manchester}{The University of Manchester, UK}

\icmlcorrespondingauthor{Petrus Mikkola}{petrus.mikkola@aalto.fi}
\icmlcorrespondingauthor{Samuel Kaski}{samuel.kaski@aalto.fi}


\vskip 0.3in
]



\printAffiliationsAndNotice{}  

\begin{abstract}
		Bayesian optimization is an effective method for finding extrema of a black-box function. We propose a new type of Bayesian optimization for learning user preferences in high-dimensional spaces. The central assumption is that the underlying objective function cannot be evaluated directly, but instead a minimizer along a projection can be queried, which we call a projective preferential query. The form of the query allows for feedback that is natural for a human to give, and which enables interaction. This is demonstrated in a user experiment in which the user feedback comes in the form of optimal position and orientation of a molecule adsorbing to a surface. We demonstrate that our framework is able to find a global minimum of a high-dimensional black-box function, which is an infeasible task for existing preferential Bayesian optimization frameworks that are based on pairwise comparisons.
\end{abstract}

\section{Introduction}

Let $f : \mathcal{X} \rightarrow \mathbb{R}$ be a \textit{black-box} function defined on a hypercube $\mathcal{X} = \prod_{d=1}^{D}[a_d,b_d]$ where $D \geq 2$. Without loss of generality we assume that $\mathbf{0} \in \mathcal{X}$. The objective is to find a global minimizer
\begin{flalign}\label{global_minimizer}
\mathbf{x}^* = \argmin_{\mathbf{x} \in \mathcal{X}} f(\mathbf{x}).
\end{flalign}
We assume, as in Preferential Bayesian Optimization \citep[PBO,][]{PBO}, that $f$ is not directly accessible. In PBO, queries to $f$ can done in pairs of points $\mathbf{x},\mathbf{x}' \in \mathcal{X}$, and the binary feedback indicates whether $f(\mathbf{x}) > f(\mathbf{x}')$. In contrast, in our work we assume that queries to $f$ are be done over the projection onto a \textit{projection vector} $\boldsymbol{\xi} \in \Xi \subset \mathcal{X}$.
The feedback is the optimal scalar projection, that is, the length $\alpha^*$ of the projection in the direction $\boldsymbol{\xi}$. We assume that there are zero coordinates in $\boldsymbol{\xi}$, and these coordinates are set to fixed values described by a \textit{reference vector} $\mathbf{x} \in \mathcal{X}$. Formally, given a query $(\boldsymbol{\xi},\mathbf{x})$, the feedback is obtained as \textit{a minimizer over the possible scalar projections},

\begin{figure}
	\begin{center}
		\includegraphics[scale=0.20]{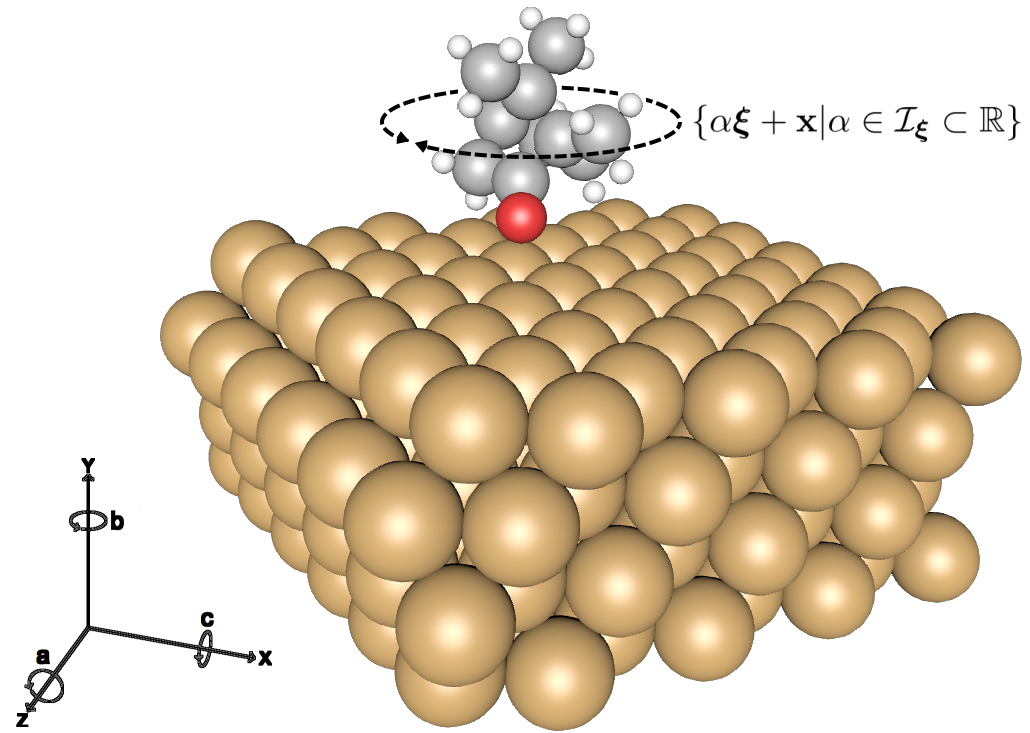}
	\end{center}
	\caption{An illustration of a projective preferential query on molecular properties: in which rotation is the molecule most likely to bind to the surface. Here, $\mathbf{x} \in \mathcal{X}$ describes the location and orientation of a molecule as a vector $\mathbf{x} = (X,Y,Z,a,b,c)$. In the figure, the projective preferential query $(\boldsymbol{\xi},\mathbf{x})$ finds the optimal rotation along the horizontal plane, defined by the coordinate $b$. This corresponds to setting $\xi_i = 0$ to all coordinates $i$ except the one corresponding to $b$, and by rotating the molecule the expert then gives the optimal value $\alpha^*$ that corresponds to the optimal value for coordinate $b$. The other coordinates are kept fixed (these are determined by $\mathbf{x}$).}\label{pp_query}
\end{figure}

\begin{flalign}\label{alpha_star}
\alpha^* = \argmin_{\alpha \in \mathcal{I}_{\boldsymbol{\xi}}} f(\alpha\boldsymbol{\xi} + \mathbf{x}),
\end{flalign}
where $\mathcal{I}_{\boldsymbol{\xi}} \equiv \{\alpha \in \mathbb{R} | \alpha\boldsymbol{\xi} + \mathbf{x} \in \mathcal{X}\}$.

What are then natural use cases for such \textit{projective preferential queries}? The main motivation comes from humans serving as the oracles. The form of the query enables efficient learning of user preferences over choice sets in which each choice has multiple attributes, and in particular over continuous choice sets. An important application is knowledge elicitation from trained professionals (doctors, physicists, etc.). For example, we may learn a material scientist’s preferences, that is, insight based on prior knowledge and experience, over molecular translations and orientations as a molecule adsorbs to a surface. In this case, a projective preferential query could correspond to finding an optimal rotation (see Figure \ref{pp_query}), which the scientist’s can easily give by rotating the molecule in a visual interface.

Probabilistic preference learning is a relatively new topic in machine learning research but has a longer history in econometrics and psychometrics \citep{mcfadden1981,mcfadden2001,Stern1990,Thurstone1927}. A wide range of applications of these models exists, for instance in computer graphics \citep{Brochu_2010}, expert knowledge elicitation \citep{AzariSoufiani_2013}, revenue management systems of airlines \citep{Carrier_2015}, rating systems, and almost any application that contains users' preference modeling. An established probabilistic model is Thurstone-Mosteller (TM) model that measures a process of pairwise comparisons \citep{Thurstone1927,Mosteller1951}. In the preference learning context, the models based on the TM-model can be applied to learning preferences from pairwise comparison feedback \citep[e.g.][]{Chu_2005}. An extension of this research into the interactive learning setting is studied by, among others, \citet{Brochu_2008} and \citet{PBO}. All these approaches resort to pairwise feedbacks. \citet{Koyama2017} proposed an extension of Bayesian optimization for learning optimal parameter values for visual design tasks by letting a user give feedback as a slider manipulation. They considered only two pairwise comparisons per slider since they tested to "included more sampling points" but they  "did not observe any significant improvement in the optimization behavior".

A drawback of most preference learning frameworks is their incapability to handle high-dimensional input spaces. The underlying reason is a \textit{combinatorial explosion} in the number of possible comparisons with respect to the number of dimensions $D$, $O(K^{2D})$, given $K$ grid-points per dimension. This implies that a single pairwise comparison has \textit{low information content}
in high-dimensional spaces. This problem was mitigated by \citet{PBO} by capturing the correlations among pairs of queries (duels). However, it is still difficult to scale that method to high-dimensional spaces, say higher than 2-dimensional (see Section \ref{numerical_experiments_section}). Furthermore, the numerical computations become infeasible in a high-dimensional setting, especially the optimization of an acquisition function or finding a Condorcet winner.  

In this paper, we introduce a Bayesian framework, which we call \textit{Projective Preferential Bayesian Optimization} (PPBO), that scales to high-dimensional input spaces. A main reason is that the information content of a projective preferential query is much higher than that of a pairwise preferential query. A projective preferential query is equivalent to infinite pairwise comparisons along a projection. An important consequence is that with projective preferential queries, the user's workload in answering the queries will be considerably reduced. 
Source code is available at \url{https://github.com/AaltoPML/PPBO}.


\section{Learning preferences from projective preferential feedback}

In this section we introduce a Bayesian framework capable of dealing with projective preferential data. A central idea is to model the user's utility function, that is $f$, as a Gaussian process as first proposed by \citet{Chu_2005}. We extend this line of study to allow projective preferential queries, by deriving a tractable likelihood, proposing a method to approximate it, and introducing four acquisition criteria for enabling interactive learning in this setting.

In this paper, for convenience, we will formulate the method for maximization instead of minimization as in (2), without loss of generality.



\subsection{Likelihood}

Our probabilistic model of user preferences is built upon the Thurstone's law of comparative judgement \citep{Thurstone1927}. A straightforward way to formalize this would be to assume pairwise comparisons are corrupted by Gaussian noise: $x \succ x'$, if and only if $f(x) + \varepsilon > f(x') + \varepsilon'$, where the latent function $f$ is \textit{a utility function} that characterizes user preferences described by the preference relation $\succ$. The standard assumption is that $\varepsilon$ and $\varepsilon'$ are identically and independently distributed Gaussians. Here, we deviate slightly from this assumption: Given two alternatives $(\alpha\boldsymbol{\xi}+\mathbf{x}), (\beta\boldsymbol{\xi}+\mathbf{x}) \in \mathcal{X}$, we assume that $\alpha\boldsymbol{\xi}+\mathbf{x} \succ \beta\boldsymbol{\xi}+\mathbf{x}$, if and only if $f(\alpha\boldsymbol{\xi}+\mathbf{x}) + W(\alpha) > f(\beta\boldsymbol{\xi}+\mathbf{x}) + W(\beta)$, where $W$ is a \textit{Gaussian white noise process} with zeromean and autocorrelation $\mathbb{E}(W(t)W(t+\tau)) = \sigma^2$ if $\tau=0$, and zero otherwise. 

We would like to find the likelihood for an observation $(\alpha,(\boldsymbol{\xi},\mathbf{x}))$ that corresponds to uncountably infinite pairwise comparisons: $\alpha\boldsymbol{\xi}+\mathbf{x} \succ \beta\boldsymbol{\xi}+\mathbf{x}$ for $\beta \neq \alpha$. For each comparison we condition on $W(\alpha)$ (more details in Supplementary material), 
\begin{flalign*}
&\textrm{P}\left(\alpha\boldsymbol{\xi}+\mathbf{x} \succ \beta\boldsymbol{\xi}+\mathbf{x}\ |\ W(\alpha) = w\right)\\
&= 1- \Phi\left(\frac{f(\beta\boldsymbol{\xi}+\mathbf{x})-f(\alpha\boldsymbol{\xi}+\mathbf{x})-w}{\sigma}\right),
\end{flalign*}
where $\Phi$ is the cumulative distribution function of the standard normal distribution. For a single comparison we have
\begin{flalign*}
&\textrm{P}\left(\alpha\boldsymbol{\xi}+\mathbf{x} \succ \beta\boldsymbol{\xi}+\mathbf{x}\right)\\
&= 1-[\Phi\ast \phi]\left(\frac{f(\beta\boldsymbol{\xi}+\mathbf{x})-f(\alpha\boldsymbol{\xi}+\mathbf{x})}{\sigma}\right),
\end{flalign*}
where $\phi$ is the probability density of the standard normal distribution and $\ast$ is the convolution operator. For infinite comparisons, we first consider a finite number of comparisons $m$. By their independence, we have
\begin{flalign*}
&\textrm{P}\left(\alpha\boldsymbol{\xi}+\mathbf{x} \succ \beta_1\boldsymbol{\xi}+\mathbf{x},...,\alpha\boldsymbol{\xi}+\mathbf{x} \succ \beta_m\boldsymbol{\xi}+\mathbf{x} \right)\\ 
&= \prod_{j=1}^{m}\left(1-[\Phi\ast \phi]\left(\frac{f(\beta_j\boldsymbol{\xi}+\mathbf{x})-f(\alpha\boldsymbol{\xi}+\mathbf{x})}{\sigma}\right)\right).
\end{flalign*}
By letting the number of points $m$ in an increasing sequence $\beta_1,...,\beta_{m}$ of the partition of the interval $\mathcal{I}_{\boldsymbol{\xi}}{\setminus}\{\alpha\}$ to approach infinity, we can interpret this as a Volterra (product) integral 
\begin{flalign*}
\exp \bigg(- \int_{\mathcal{I}_{\boldsymbol{\xi}}} [\Phi \ast \phi]\Big(\frac{f(\beta\boldsymbol{\xi}+\mathbf{x}) - f(\alpha\boldsymbol{\xi}+\mathbf{x})}{\sigma}\Big)d\beta \bigg).
\end{flalign*}
The joint log-likelihood of a dataset $\mathcal{D}$, denoted as $\mathcal{L}(\mathcal{D} | \mathbf{f})$, takes the form
\begin{flalign*}
-\sum_{i=1}^{N} \int_{\mathcal{I}_{\boldsymbol{\xi}^i}} [\Phi \ast \phi] \Big(\frac{f(\beta\boldsymbol{\xi}^i+\mathbf{x}^i) - f(\alpha^i\boldsymbol{\xi}^i+\mathbf{x}^i)}{\sigma}\Big)d\beta.
\end{flalign*}

\subsection{Prior}

First, we introduce notation. Assume that $N$ projective preferential queries have been performed and gathered into a dataset $\mathcal{D} = \{(\alpha^{i},(\boldsymbol{\xi}^{i},\mathbf{x}^{i}))\}_{i=1}^N$. For every data instance $(\alpha^{i},(\boldsymbol{\xi}^{i},\mathbf{x}^{i}))$, we also consider a sequence of pseudo-observations $\{(\beta_j^i\boldsymbol{\xi}^{i},\mathbf{x}^{i})\}_{j=1}^{m}$. Technically, the pseudo-observations are Monte-Carlo samples needed for integrating the likelihood. The latent function values evaluated on those points are gathered into a vector,
\begin{flalign*}
\mathbf{f}^{(i)} \equiv \big(f(\alpha^{i}\boldsymbol{\xi}^{i}+\mathbf{x}^{i}),\{f(\beta_j^i\boldsymbol{\xi}^{i}+\mathbf{x}^{i})\}_{j=1}^{m}\big).
\end{flalign*}
The latent function vector over all points is formed by concatenating over $\mathbf{f} \equiv (\mathbf{f}^{(i)})_{i=1}^N$. 

The user's utility function $f$ is modelled as a Gaussian process \cite{Williams_Rasmussen}. GP model fits ideally to this objective, since it is flexible (non-parametric) and can conveniently handle uncertainty (predictive distributions can be derived analytically). In particular, it allows us to have insight into those regions of the space $\mathcal{X}$ in which either we are uncertain about user preferences due to lack of data, or because the user gives inconsistent feedback. A possible reason for the latter is that one of the preference axioms, \textit{transitivity} or \textit{completeness}, is violated in those regions. A weak preference relation $\succeq$ is \textit{complete} if for all $\mathbf{x},\mathbf{y} \in \mathcal{X}$, either $\mathbf{x} \succeq \mathbf{y}$ or $\mathbf{y} \succeq \mathbf{x}$ holds. That is, a user is able to reveal their preferences over all possible pairwise comparisons. Similarly, $\succeq$ is \textit{transitive} if for any $\mathbf{x},\mathbf{y},\mathbf{z} \in \mathcal{X}$ the following holds: $(\mathbf{x} \succeq \mathbf{y}$ and $\mathbf{y} \succeq \mathbf{z})$ implies that $\mathbf{x} \succeq \mathbf{z}$. That is, a user has consistent preferences. This together with the continuity of $\succeq$ guarantees the existence of a real-valued continuous utility function that represents $\succeq$ \cite{Debreu1954}.

Thus, we assume as \citet{Chu_2005}, that the prior of the utility function follows a zero-mean Gaussian process,
\begin{flalign*}
p(\mathbf{f}) = \frac{1}{(2\pi)^{\frac{N}{2}}|\Sigma|^{\frac{1}{2}}} \exp(-\frac{1}{2}\mathbf{f}^{\top}\Sigma^{-1}\mathbf{f}),
\end{flalign*}
where the $ij^{th}$-element of the covariance matrix is determined by a kernel $k$ as $\Sigma_{ij} = k(\mathbf{x}^i,\mathbf{x}^j)$. Throughout the paper, we assume the \textit{squared exponential} kernel $k(\mathbf{x},\mathbf{x}) = \sigma_f^2 \exp(\frac{1}{-2l}\norm{\mathbf{x}-\mathbf{x}'}^2)$, where the $\sigma_f$ and $l$ are hyperparameters. 

\subsection{Posterior}

For the sake of simplicity, we use the Laplace approximation for the posterior distribution. A maximum a posteriori (MAP) estimate is needed for that,
\begin{flalign*}
&\argmax_{\mathbf{f}}\textrm{P}(\mathbf{f}|\mathcal{D})= \argmax_{\mathbf{f}}( \textrm{P}(\mathbf{f})\mathcal{L}(\mathcal{D}|\mathbf{f})) = \argmax_{\mathbf{f}}T(\mathbf{f}),
\end{flalign*}
where we denote the functional (log-scaled posterior)
\begin{flalign*}
&T(\mathbf{f}) \equiv -\frac{1}{2}\mathbf{f}^{\top}\Sigma^{-1}\mathbf{f}&\\
&- \sum_{i=1}^{N} \int_{\mathcal{I}_{\boldsymbol{\xi}^i}} [\Phi \ast \phi] \Big(\frac{f(\beta\boldsymbol{\xi}^i+\mathbf{x}^i) - f(\alpha^i\boldsymbol{\xi}^i+\mathbf{x}^i)}{\sigma}\Big)d\beta.&
\end{flalign*}
The convolution $\Phi \ast \phi$ can be efficiently approximated by Gauss-Hermite quadrature. The outer integral is approximated as a Monte-Carlo integral,
\begin{flalign*}
&\int_{\mathcal{I}_{\boldsymbol{\xi}}} [\Phi \ast \phi] \Big(\frac{f(\beta\boldsymbol{\xi}+\mathbf{x}) - f(\alpha\boldsymbol{\xi}+\mathbf{x})}{\sigma}\Big)d\beta&\\
&\approx \frac{\ell(\mathcal{I}_{\boldsymbol{\xi}})}{m}\sum_{j=1}^{m}[\Phi \ast \phi] \Big(\frac{f(\beta_j\boldsymbol{\xi}+\mathbf{x}) - f(\alpha\boldsymbol{\xi}+\mathbf{x})}{\sigma}\Big),&
\end{flalign*}
where the pseudo-observations $(\beta_j\boldsymbol{\xi})_{j}^{m}$ for $j=1,...,m$ are sampled from a suitable distribution.
Our choice is to use a family of truncated generalized normal (TGN) distributions, since it provides a continuous transformation from the uniform distribution to the truncated normal distribution, such that the locations of distributions can be specified. The idea is to concentrate pseudo-observations more densely around the optimal value $\alpha\boldsymbol{\xi}$ as the number of queries increases. For more details, see Supplementary material. 

For notational convenience, define 
\begin{flalign*}
\Delta_{i,j}(\mathbf{f}) \equiv \frac{f(\beta_j^i\boldsymbol{\xi}^i+\mathbf{x}^i) - f(\alpha^i\boldsymbol{\xi}^i+\mathbf{x}^i)}{\sigma}.
\end{flalign*}
If the domain is normalized to $\mathcal{X} = \prod_{d=1}^{D}[0,1]$, and the projections are normalized to $\xi = \frac{\xi}{\norm{\xi}_{\infty}}$, then $\ell(\mathcal{I}_{\boldsymbol{\xi}})=1$. Hence, under this normalization, the functional $T$ can be approximated as
\begin{flalign*}
T(\mathbf{f}) \approx -\frac{1}{2}\mathbf{f}^{\top}\Sigma^{-1}\mathbf{f} - \frac{1}{m}\sum_{i=1}^{N}\sum_{j=1}^{m}[\Phi \ast \phi] \big(\Delta_{i,j}(\mathbf{f})\big).
\end{flalign*} 
The MAP estimate can be efficiently solved by a second-order iterative optimization algorithm, since the gradient and the Hessian can be easily derived for $T$.

The Laplace approximation of the posterior amounts to the second-order Taylor approximation of the log posterior around the MAP estimate. In the ordinary (non-log) scale, this reads
\begin{flalign*}
\textrm{P}(\mathbf{f}|\mathcal{D}) \approx \textrm{P}(\mathbf{f}_{\textrm{MAP}}|\mathcal{D})\exp\big(-\frac{1}{2}(\mathbf{f}-\mathbf{f}_{\textrm{MAP}})^{\top} \textrm{H} (\mathbf{f}-\mathbf{f}_{\textrm{MAP}})\big),
\end{flalign*}
where the matrix $\textrm{H}$ is the negative Hessian of the log-posterior at the MAP estimate, $\textrm{H}\equiv-\nabla\nabla\log \textrm{P}(\mathbf{f}|\mathcal{D})|_{\mathbf{f}=\mathbf{f}_{\textrm{MAP}}}= \Sigma^{-1}+\Lambda$.\footnote{We denote the partial derivatives matrix evaluated at MAP estimate as \begin{flalign*}
	\Lambda \equiv \frac{\partial^2}{\partial \mathbf{f} \partial \mathbf{f}^{\top}}\frac{1}{m}\sum_{i=1}^{N} \sum_{j=1}^{m}[\Phi \ast \phi] \big(\Delta_{i,j}(\mathbf{f})\big) \bigg|_{\mathbf{f}=\mathbf{f}_{\textrm{MAP}}}.
	\end{flalign*}} In other words, the posterior distribution is approximated as a multivariate normal distribution with mean $\mathbf{f}_{\textrm{MAP}}$ and the covariance matrix $(\Sigma^{-1}+\Lambda)^{-1}$.

\subsection{Predictive distribution}

Based on the well-known properties of the multivariate Gaussian distribution, the predictive distribution of $\mathbf{f}$ is also Gaussian. Given test locations $(\mathbf{y}^{(1)},...,\mathbf{y}^{(M)})$, consider the $N$ by $M$ matrix $K\equiv [k(\mathbf{y}^{(j)},\mathbf{x}^{(i)})]_{ij}$. The predictive mean and the predictive covariance at test locations are (for more details see \cite{Williams_Rasmussen} or \cite{Chu_2005})
\begin{flalign*}
\boldsymbol{\mu}_{\textrm{pred}} &= K^{\top}\Sigma^{-1}\mathbf{f}_{\textrm{MAP}}&\\
\Sigma_{\textrm{pred}} &= \Sigma' - K^{\top}(\Sigma + \Lambda^{-1})^{-1} K,&
\end{flalign*}       
where $\Sigma'$ is the covariance matrix of the test locations.    

\section{Sequential learning by projective preferential query}\label{sequential_learning}

In this section, we discuss how to select the next projective preferential query $(\boldsymbol{\xi},\mathbf{x})$. We will choose the next query as a maximizer of an \textit{acquisition function} $\alpha(\boldsymbol{\xi},\mathbf{x})$, for instance, we will consider a modified version of the \textit{expected improvement} acquisition function \cite{Jones1998}. The optimization $(\boldsymbol{\xi},\mathbf{x})_{next} = \argmax_{(\boldsymbol{\xi},\mathbf{x})}\alpha(\boldsymbol{\xi},\mathbf{x})$ is carried out by using Bayesian optimization (more details in Supplementary material).  

If the oracle is a human, this allows us to learn user preferences in an iterative loop, making PPBO \textit{interactive}. However, this interesting special case, where $f$ is a utility function of a human, also brings forth issues due to \textit{bounded rationality}. We apply here the following narrow definition of this more general concept \citep[see][]{simon1990reason}: Bounded rationality is the idea that users give feedback that reflects their preferences, but within the limits of the information available to them and their mental capabilities.


\subsection{The effects of bounded rationality on the optimal next query}\label{bounded_rationality_section}

If the oracle is a human, it is important to realize that the optimal next $(\boldsymbol{\xi},\mathbf{x})$ is not solely the one which optimally balances the \textit{exploration-exploitation trade-off} -- as it is for a perfect oracle -- but the optimal $(\boldsymbol{\xi},\mathbf{x})$ takes also into account human cognitive capabilities and limitations. For instance, the more there are non-zero coordinates in $\boldsymbol{\xi}$, the greater the "cognitive burden" to a human user, and the harder it becomes to give useful feedback. Thus, if there is a human in the loop, the choice of $\boldsymbol{\xi}$ should take into account both the optimization needs and what types of queries are convenient for the user.

The projective preferential feedback \eqref{alpha_star} may not be single-valued or even well-defined for all $\boldsymbol{\xi} \in \Xi$, if the oracle is a human. For instance, the user may not be able to explicate their preferences with respect to the $d^{th}$ attribute, that is, the preferences do not satisfy the \textit{completeness axiom}. Formally, this means that if $\boldsymbol{\xi}=\mathbf{e}_d$ (the $d^{th}$-standard unit vector), then for some $\mathbf{x} \in \mathcal{X}$ it holds that $\argmax_{\alpha \in \mathcal{I}_{\boldsymbol{\xi}}} f(\alpha \boldsymbol{\xi} + \mathbf{x})$ is multi-valued, a random variable or not well-defined -- depending on how we interpret the scenario in which the user should say "I do not know" but instead gives an arbitrary feedback. Fortunately, this incompleteness can be easily handled when a GP is used for modelling $f$; it just implies that the posterior variance is high along the dimension $d$. 

A possible solution would be to allow the answer "I do not know", and to design an acquisition function that is capable of discovering and avoiding those regions in the space $\mathcal{X}$ where the user gives inconsistent feedback due to any source of bounded rationality. This challenge is left for future research. It is noteworthy that the acquisition function we introduce next, performed well in the user experiment covered in Section \ref{user_experiment}.

\subsection{Expected improvement by projective preferential query}

We define \textit{the expected improvement by projective preferential query} at the $n^{th}$-iteration by
\begin{flalign}\label{expected_improvement}
\textrm{EI}_n (\boldsymbol{\xi},\mathbf{x}) \equiv \mathbb{E}_n \Big(\max \big\{\max_{\alpha \in \mathcal{I}_{\boldsymbol{\xi}}}f(\alpha\boldsymbol{\xi}+\mathbf{x}) - \mu^{*}_n, 0 \big\} \Big),
\end{flalign}
where $\mu^{*}_n$ denotes the highest value of the predictive posterior mean, and the expectation is conditioned on the data up to the $n^{th}$-iteration. The maximum over $\alpha$ models the anticipated feedback.

$\textrm{EI}_n$ can be approximated as a Monte-Carlo integral (up to a multiplicative constant that does not depend on $(\mathbf{\boldsymbol{\xi}},\mathbf{x})$),
\begin{flalign}\label{expected_improvement_MC}
\frac{1}{K}\sum_{k=1}^{K} \max \big\{\max_{\alpha \in \mathcal{I}_{\boldsymbol{\xi}}}\tilde{f}_k(\alpha\boldsymbol{\xi}+\mathbf{x}) - \mu^{*}_n, 0 \big\},
\end{flalign}
where $\max_{\alpha \in \mathcal{I}_{\boldsymbol{\xi}}}\tilde{f}_k(\alpha\boldsymbol{\xi}+\mathbf{x})$ is approximated by using \textit{discrete}\footnote{Another alternative is to consider \textit{continuous Thompson sampling} to draw a continuous sample path from the GP model, and then maximize it. The method is based on Bochner's theorem and the equivalence between a Bayesian linear model with random features and a Gaussian process. For more details see \cite{HernandezLobato2014}.} \textit{Thompson sampling} as described in \cite{HernandezLobato2014}. Discrete Thompson sampling draws a finite sample from the GP posterior distribution, and then returns the maximum over the sample. The steps needed to approximate $\textrm{EI}_n$ are summarized in Algorithm \ref{EI_algorithm}. 
\begin{algorithm}
	\caption{Approximate $\textrm{EI}_n(\boldsymbol{\xi},\mathbf{x})$}
	\label{EI_algorithm}
	\begin{algorithmic}
	    \INPUT $(\boldsymbol{\xi},\mathbf{x})$ and $K \geq 1$, $J \geq 1$
		\STATE 1. Compute $(\Sigma + \Lambda^{-1})^{-1}$
		\FOR{$k = 1,2,...,K$}
			\STATE 2. Draw $(\beta_j)_{j=1}^{J}$\\
			\STATE 3. Draw $\big(\tilde{f}(\beta_j\boldsymbol{\xi} + \mathbf{x})\big)_{j=1}^{J}$ from the predictive distribution $\mathcal{N}\big(\boldsymbol{\mu}_{\textrm{pred}},\Sigma_{\textrm{pred}}\big)$ \\
			\STATE 4. $z_k \leftarrow \max_j \big\{ \big(\tilde{f}(\beta_j\boldsymbol{\xi} + \mathbf{x})\big)_{j=1}^{J} \big\}$\\
		\ENDFOR \\
		\OUTPUT $\frac{1}{K}\sum_{k=1}^{K}\max\big\{z_k-\mu^{*}_n,0\big\}$
	\end{algorithmic}
\end{algorithm}

The bottlenecks are the first and the third steps. In the third step, a predictive covariance matrix of size $J\times J$ needs to be computed, and then a sample from the multivariate normal distribution needs to be drawn. Hence, the time complexity of Algorithm \ref{EI_algorithm} is 
$O(N^3m^3 + KN^2m^2J + KNmJ^2 + KJ^3)$,
where the terms come from a matrix inversion (the first step), two matrix multiplications, and a Cholesky decomposition, respectively. Recall that $N,m,K$ and $J$ refer to the number of observations, pseudo-observations, Monte-Carlo samples, and grid points, respectively.

\subsection{Pure exploitation and exploration}
In the experiments we use pure exploration and exploitation as baselines. A natural interpretation of pure exploitation in our context is to select the next query $(\boldsymbol{\xi},\mathbf{x})$ such that $\boldsymbol{\xi} + \mathbf{x} = \mathbf{x}^* \equiv \argmax_{\mathbf{x}' \in \mathcal{X}}\mu_n(\mathbf{x}')$, where $\mu_n(\mathbf{x}')$ is the posterior mean of the GP model at location $\mathbf{x}'$, given all the data so far $\mathcal{D}_n$. 

We interpret pure exploration as maximization of the GP variance on a query $(\boldsymbol{\xi},\mathbf{x})$ given the anticipated feedback. That is, the pure explorative acquisition strategy maximizes the following acquisition function
\begin{flalign}\label{varince_max}
\textrm{Explore}(\boldsymbol{\xi},\mathbf{x}) \equiv \mathbb{V}_n\Big(\max_{\alpha \in \mathcal{I}_{\boldsymbol{\xi}}}f(\alpha\boldsymbol{\xi}+\mathbf{x}) \Big).
\end{flalign}
In practice, \eqref{varince_max} is approximated by Monte-Carlo integration and discrete Thompson sampling in the same vein as in Algorithm \ref{EI_algorithm}.

\subsection{Preferential coordinate descent (PCD)}

The fourth acquisition strategy corresponds to the interesting special case where $\boldsymbol{\xi} = \mathbf{e}_d$ (the $d^{th}$-standard unit vector), and the coordinates $d$ are rotated in a cyclical order for each query. The reference vector $\mathbf{x} = (x_1,...,x_{d-1},0,x_{d+1},...,x_D)$ can be chosen in several ways, but it is natural to consider an exploitative strategy in which $\mathbf{x}$ is set to $\mathbf{x}^*$ except for the $d^{th}$-coordinate which is set to zero. We call this acquisition strategy \textit{Preferential Coordinate Descent} (PCD), since PPBO with PCD acquisition is closely related to a \textit{coordinate descent algorithm} that successively minimizes an objective function along coordinate directions. The PPBO method with PCD acquisition (PPBO-PCD) differs from the classical coordinate descent in two ways: First, PPBO-PCD assumes that direct function evaluations are not possible but instead projective preferential queries are. Second, it models the black-box function $f$ (as a GP) whereas the classical coordinate descent does not. This makes PPBO-PCD able to take advantage of past queries from every one-dimensional optimization. 

When comparing to other acquisition strategies, we show that PCD performs well in numerical experiments (when $f$ is not a utility function but a numerical test function). This agrees with the results in the optimization literature; for instance: if $f$ is pseudoconvex with continuous gradient, and $\mathcal{X}$ is compact and convex with "nice boundary", then the coordinate descent algorithm converges to a global minimum \citep[Corollary 3.1]{Spall2012}. However, PCD may not perform so well in high-dimensional spaces, since it cannot query in between the dimensions. For instance, the expected improvement by projective preferential query outperformed PCD on a 20D test function (see Section \ref{numerical_experiments_section}), since it allows to query arbitrary projections.  

\section{Numerical experiments}\label{numerical_experiments_section}
In this section we demonstrate the efficiency of the PPBO method in high-dimensional spaces, and experiment with various acquisition strategies in numerical experiments on simulated functions. 

The goal is to find a global minimum of a black-box function $f$ by querying it either through (i) pairwise comparisons or (ii) projective preferential queries. For (i) we use the PBO method of \citet{PBO}, which is state of the art among Gaussian process preference learning frameworks that are based on pairwise comparisons. For (ii) we use the PPBO method as introduced in this paper. The four different acquisition strategies introduced in Section \ref{sequential_learning} are compared against the baseline that samples a random $(\boldsymbol{\xi},\mathbf{x})$. For the PBO method, we consider a random and a dueling-Thompson sampling acquisition strategies. In total seven different methods are compared: the expected improvement by projective preferential query (\textsc{ppbo-ei}), pure exploitation (\textsc{ppbo-ext}), pure exploration (\textsc{ppbo-exr}), preferential coordinate descent (\textsc{ppbo-pcd}), random (\textsc{ppbo-rand}), and for the PBO; random (\textsc{pbo-rand}) and a variant of dueling-Thompson sampling (\textsc{pbo-dts}). For more details, see Supplementary material.

\begin{figure*}
	\centering
	\includegraphics[scale=0.145]{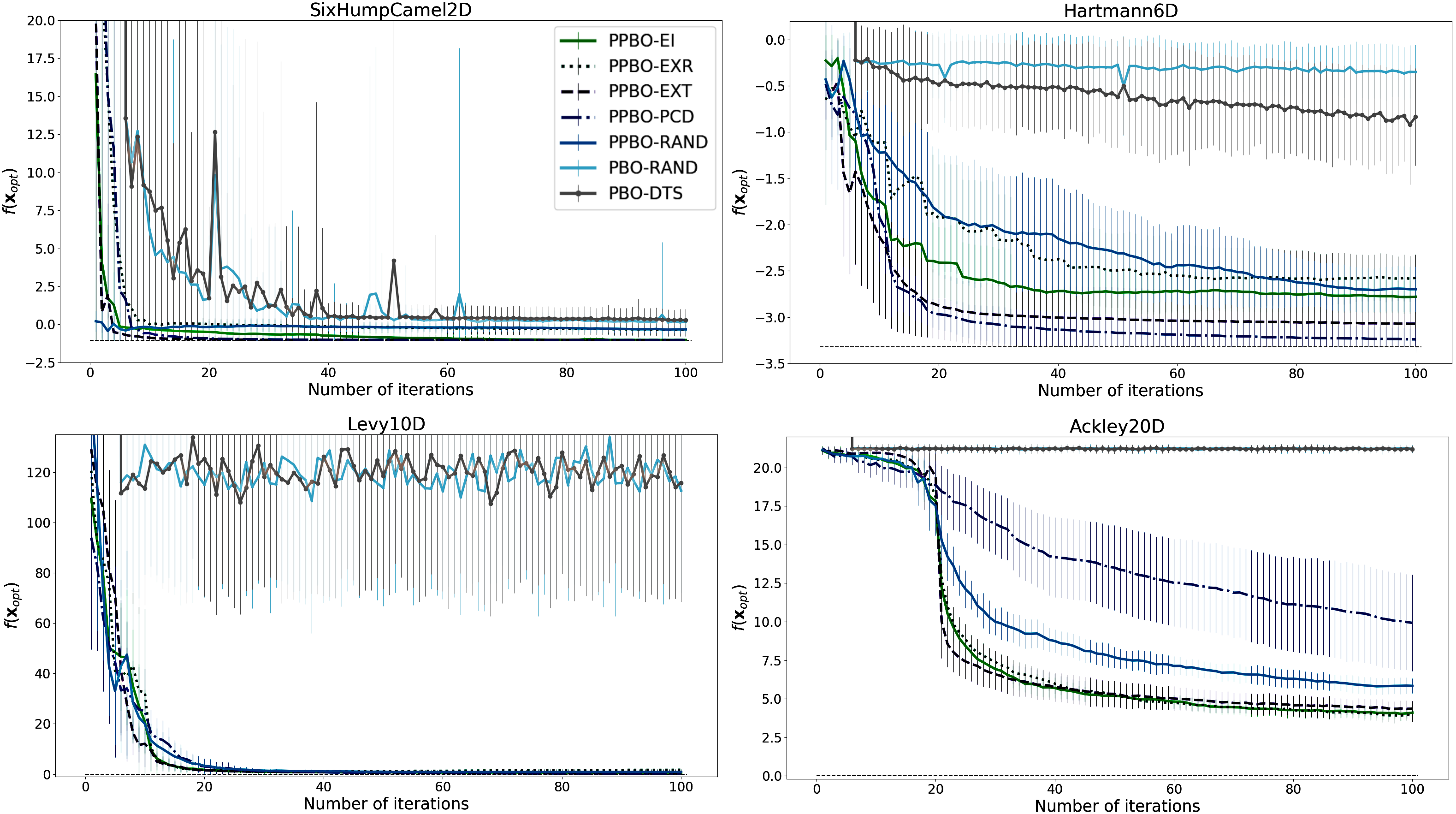}
	\caption{Convergence of different methods across 4 objective functions. The results are averaged over 25 different random initializations (for PBO we run 100 seeds). The vertical axis represent the value of the true objective function at the best guess: $\mathbf{x}_{opt} = \argmax_{\mathbf{x}}\mu_n(\mathbf{x})$ in PPBO, and $\mathbf{x}_{opt} = "Condorcet\ winner"$ in PBO. The horizontal axis represents the number of projective preferential queries in PPBO, and the number of pairwise comparisons in PBO. The black horizontal dashed line indicates the global minimum of the objective function (a small Gaussian noise is added to the function values). The standard deviations are depicted by the vertical lines.}\label{fig_testfunctions}
\end{figure*}

For $f$ we consider four different test functions: Six-hump-camel2D, Hartmann6D, Levy10D and Ackley20D.\footnote{
https://www.sfu.ca/$\sim$ssurjano/optimization.html} We add a small Gaussian error term to the test function outputs. There are as many initial queries as there are dimensions in a test function. The $i^{th}$-initial query corresponds to $\boldsymbol{\xi} = \mathbf{e}_i$, that is, to the $i^{th}$-coordinate projection, and the reference vector $\mathbf{x}$ is uniformly random. We consider a total budget of 100 queries. The results are depicted in Figure \ref{fig_testfunctions}.\footnote{All experiments of each test function were run on a computing infrastructure of 24x Xeon Gold 6148 2.40GHz cores and 72GB RAM. The longest experiment (Ackley20D) took in total 24h.}

\textsc{ppbo-pcd} obtained the best performance on three of the four test functions. On the high-dimensional test function Ackley20D, \textsc{ppbo-ei} performed best. Unsurprisingly, all PPBO variants clearly outperformed all PBO variants. Since the performance gap between \textsc{ppbo-rand} and \textsc{pbo-rand} is so high, we conclude that from the optimization perspective; whenever a projective preferential query is possible, a PPBO-type of approach should be preferred to an approach that is based on pairwise comparisons. However, we note that it is better to think of PPBO as a complement, not as a substitute for PBO. In the applications, pairwise comparisons may be preferred, for instance, if they are more convenient to a user, and the underlying choice space is low-dimensional.  

To illustrate the low information content of pairwise comparisons, we ran a test on the Six-hump-camel2D function. We trained a GP classifier with 2000 random queries (duels), and found a Condorcet winner \citep[see][]{PBO} by maximizing the soft-Copeland score ($33\times33$ MC-samples used for the integration) by using Bayesian optimization (500 iterations with 10 optimization restarts). This took 41 minutes on the $8^{th}$-gen Intel i5-CPU, and the distance to a true global minimizer was $\norm{\mathbf{x}_{c} - \mathbf{x}_{true}} = \norm{(0.1770,-0.0488)-(0.0898,-0.7126)}\approx0.67$, and the corresponding function value was $0.1052$ compared to a true global minimum value $-1.0316$. In contrast, \textsc{ppbo-rand} reached this level of accuracy at the first queries, as seen from Figure \ref{fig_testfunctions}.


\section{User experiment}\label{user_experiment}
In this section we demonstrate the capability of PPBO to correctly and efficiently encode user preferences from projective preferential feedback. 

We consider a material science problem of a single organic molecule adsorbing to an inorganic surface. This is a key step in understanding the structure at the interface between organic and inorganic films inside electronic devices, coatings, solar cells and other materials of technological relevance. The molecule can bind in different adsorption configurations, altering the electronic properties at the interface and affecting device performance. Exploring the structure and property phase space of materials with accurate but costly computer simulations is a difficult task. Our objective is to find the most stable surface adsorption configuration through human intuition and subsequent computer simulations. The optimal configuration is the one that minimises the computed adsorption energy.

Our test case is the adsorption of a non-symmetric, bulky molecule camphor on the flat surface of (111)-plane terminated Cu slab. Some understanding of chemical bonding is required to infer correct adsorption configurations. The user is asked to consider the adsorption structure as a function of molecular orientation and translation near the surface. These are represented with 6 physical variables: angles $\alpha$, $\beta$, $\gamma$ of molecular rotation around the X, Y, Z Cartesian axes (in the range [0, 360] deg.), and distances x, y, z of translation above the surface (with lattice vectors following the translational symmetry of the surface). The internal structures of the molecule and surface were kept fixed since little structural deformation is expected with adsorption. A similar organic/inorganic model system and experiment scenario was previously employed to detect the most stable surface structures with autonomous BO, given the energies of sampled configurations \citep{BOSS}.

In this interactive experiment, the users encode their preferred adsorption geometry as a location in the 6-dimensional phase space. We employ the quantum-mechanical atomistic simulation code FHI-aims \citep{FHI-AIMS} to $i$) compute the adsorption energy \textbf{E} of this preferred choice, and $ii$) optimise the structure from this initial position to find the nearest local energy minimum in phase space, \textbf{E*}. We also consider the number of optimization steps \textbf{N} needed to reach the nearest minimum as a measure of quality of the initial location.

There are four different test users: two materials science experts (human: a PhD student and an experienced researcher, both of them know the optimal solution), a non-expert (human), and a random bot (computer). The hypothesis is that: \textit{the materials science experts should obtain structures associated with lower energy minimum points}. We consider only coordinate projections, that is $\boldsymbol{\xi} \in \{\mathbf{e}_1,...,\mathbf{e}_6\}$. In other words, we let the user choose the optimal value for one dimension at a time.

The total number of queries was 24, of which 6 were initial queries. The $i^{th}$-initial query corresponded to $\boldsymbol{\xi} = \mathbf{e}_i$, that is, to the $i^{th}$-coordinate projection. The initial values for the reference coordinate vector $\mathbf{x}$ were fixed to the same value across all user sessions. For acquisition, we used the expected improvement by projective preferential query. Since we allowed only coordinate projections for $\boldsymbol{\xi}$, we first selected $\boldsymbol{\xi}_{n+1} = \argmax_{\boldsymbol{\xi} \in \{\mathbf{e}_1,...,\mathbf{e}_6\} }\int \textrm{EI}_{n}(\boldsymbol{\xi},\mathbf{x})d\mathbf{x}$, and then, either $\mathbf{x}_{n+1}=\argmax_{\mathbf{x}} \mu_n(\mathbf{x})$ (\textsc{ei-ext}), or the next $\mathbf{x}_{n+1}$ was drawn uniformly at random (\textsc{ei-rand}). The computer bot gave random values to the queries; to  provide some consistency to the bot, $\mathbf{x}_{n+1}$ was selected by maximizing a standard expected improvement function (\textsc{ei-ei}). The results are summarized in Table \ref{results-table}.    

\begin{table}
	\caption{Results of the user experiment. The absorption energies were computed by using density functional theory (DFT) methods. The energies represent the absorption energies of the relaxed structures corresponding to the most preferred configurations.}\label{results-table}
	\begin{center}
		\begin{tabular}{l c r r r}
			\textbf{User}  &\textbf{Acq. of} $(\boldsymbol{\xi},\mathbf{x})$ & \textbf{E (eV)} & \textbf{E* (eV)} & \textbf{N} \\
			\hline \\
			Expert 1         &\textsc{ei-rand} &-0.454  &-1.023 &  62 \\
			Expert 1        &\textsc{ei-ext}  &-0.371 &-1.029 & 39 \\
			Expert 2         &\textsc{ei-rand}  &-0.611 &-1.007 &  37 \\
			Expert 2         &\textsc{ei-ext}  &-0.564 &-1.030 & 45 \\
			Non-expert         &\textsc{ei-rand}  &-0.481 &-0.771 & 32 \\
			Non-expert         &\textsc{ei-ext}  &-0.511 &-0.762 & 56 \\
			Bot         &\textsc{ei-ei}  &-0.365 &-0.643 & 70 \\
			Bot         &\textsc{ei-ei}  &-0.612 &-0.753 & 63 \\
			Bot         &\textsc{ei-ei}  &2.231 &-0.959 & 127 \\
			Bot         &\textsc{ei-ei}  &-0.216 &-0.783 & 88
		\end{tabular}
	\end{center}
\end{table}

Our first observation is that PPBO can distinguish between the choices made by a human and a computer bot. Human choices pinpoint atomic arrangements that are close to nearby local minima (small \textbf{N}), while the random bot's choices are far less reasonable and require much subsequent computation to optimise structures. For all human users, the preferred molecular structures were placed somewhat high above the surface, which led to relatively high \textbf{E} values. With this query arrangement, it appears the z variable was the most difficult one to estimate visually. Human-preferred molecular orientations were favourable, so the structures were optimised quickly (few \textbf{N} steps).    

The quality of user preference is best judged by the depth of the nearest energy basin, denoted by \textbf{E*}. It describes the energy of the structure preferred by a user. Here, there is a marked divide by expertise. The structures refined from the choices of the bot and non-expert are local minima of adsorption, characterised by weak dispersive interactions. The expert's choice led to two low-energy structure types that compete for the global minimum, and feature strong chemical bonding of the O atom to the Cu surface. Thus, the data (Table \ref{results-table}, column \textbf{E*}: rows 1-4 versus rows 5-10) supports our hypothesis: the materials science experts do obtain structures associated with lower energy minimum points. 

The findings above demonstrate that the PPBO framework was able to encode the expert knowledge described via preferences. However, since there are only 10 samples, further work will be needed to validate the results.


\section{Conclusions}
In this paper we have introduced a new Bayesian framework, PPBO, for learning user preferences from a special kind of feedback, which we call projective preferential feedback. The feedback is equivalent to a minimizer along a projection. Its form is especially applicable in a \textit{human-in-the-loop} context. We demonstrated this in a user experiment in which the user gives the feedback as an optimal position or orientation of a molecule adsorbing to a surface. PPBO was capable of encoding user preferences in this case. 

We demonstrated that PPBO can deal with high-dimensional spaces where existing preferential Bayesian optimization frameworks that are based on pairwise comparisons, such as IBO \citep{Brochu_2010_thesis} or PBO \citep{PBO}, have difficulties to operate. In the numerical experiments, the performance gap between PPBO and PBO was so high that we conclude: whenever a projective preferential query is possible, a PPBO-type of approach is preferable from the optimization perspective. However, we note that it is better to think of PPBO as a complement, not as a substitute for PBO. In the applications, pairwise comparisons may be preferred, for instance, if they are more convenient to a user.

In summary, if it is possible to query a projective preferential query, then PPBO provides an efficient way for preference learning in high-dimensional problems. In particular, PPBO can be used for efficient expert knowledge elicitation in high-dimensional settings which are important in many fields.

\section*{Acknowledgements}
This research was supported by the Academy of Finland (Flagship programme: Finnish Center for Artificial Intelligence FCAI; and grants 320181, 319264, 313195, 292334 and 316601). Computational resources were provided by the CSC IT Center for Science, and the Aalto Science-IT Project.

\bibliography{Projective_Preferential_Bayesian_Optimization}
\bibliographystyle{icml2020}

\end{document}